\documentclass[letterpaper, 10 pt, conference]{ieeeconf}  

\IEEEoverridecommandlockouts                              

\usepackage{graphics} 
\usepackage{booktabs}
\usepackage{epsfig} 
\usepackage{times} 
\usepackage{amsmath} 
\usepackage{amssymb}  
\usepackage{mathtools}
\usepackage{tikz}
\usepackage{standalone}
\usepackage{bondgraphs}
\usepackage{url}
\usepackage{blox}
\usepackage{gensymb}
\let\temp\rmdefault
\usepackage{mathpazo}
\let\rmdefault\temp
\usepackage{tabularx}
\usepackage{amsfonts,dcolumn}
\usepackage{blindtext}
\usepackage{rotating}
\usepackage{color}
\usepackage{transparent}
\usepackage{algorithm}
\usepackage{float}
\usepackage{algorithmic}
\usepackage{balance}
\usepackage{upgreek}
\usepackage[miama]{emf}
\DeclareMathAlphabet{\mathcal}{OMS}{cmsy}{m}{n}

\usetikzlibrary{shapes,arrows}
\usetikzlibrary{calc, patterns, decorations.pathmorphing, decorations.markings, shapes.geometric, arrows, positioning, angles, automata, positioning}
\usepackage{cleveref}

\usepackage[keeplastbox]{flushend}
\usepackage{balance}

\usepackage{textcomp}
\usepackage{lipsum}
\usepackage{cuted}
\allowdisplaybreaks
\usepackage{accents}

\newcommand{\R}{\mathbb{R}}
\newcommand{\U}{\large{\texttt{U}}}

\DeclarePairedDelimiter{\norm}{\lVert}{\rVert}
\DeclarePairedDelimiter{\abs}{\lvert}{\rvert}%

\mathchardef\mhyphen="2D
\mathchardef\mslash ="202F

\usepackage{graphicx} 
\usepackage{amsmath} 
\usepackage{amssymb}  

\title{\LARGE \bf BP-RRT: Barrier Pair Synthesis for Temporal Logic Motion Planning}
\author{Binghan He$^{1}$, Jaemin Lee, Ufuk Topcu and Luis Sentis
\thanks{
This work was supported by the National Science Foundation \texttt{\small [grant number 1724360]} and the Office of Naval Research \texttt{\small [grant number N000141512507]}.
The authors are with the Department of Mechanical Engineering \texttt{\small(B.H., J.L.)} and the Department of Aerospace Engineering and Engineering Mechanics \texttt{\small(U.T., L.S.)}, The University of Texas at Austin, Austin, TX.
Send correspondence to $^{1}$$\;${\tt\small binghan at utexas dot edu}.
}
}

\newcommand\copyrighttext{%
  \scriptsize 
  Accepted for publication in IEEE Conference on Decision and Control (CDC)
  \textcopyright 2020 IEEE. Personal use of this material is permitted. Permission from IEEE must be obtained for all other uses, in any current or future media, including reprinting/republishing this material for advertising or promotional purposes, creating new collective works, for resale or redistribution to servers or lists, or reuse of any copyrighted component of this work in other works.
  }
\newcommand\copyrightnotice{%
\begin{tikzpicture}[remember picture,overlay]
\node[anchor=south,yshift=10pt] at (current page.south)
{\fbox{\parbox{\dimexpr\textwidth-\fboxsep-\fboxrule\relax}{\copyrighttext}}};
\end{tikzpicture}%
}

\begin{document}
\maketitle
\thispagestyle{empty}
\pagestyle{empty}
\copyrightnotice

\vspace{-10pt}
\begin{abstract}
For a nonlinear system (e.g. a robot) with its continuous state space trajectories constrained by a linear temporal logic specification, the synthesis of a low-level controller for mission execution often results in a non-convex optimization problem. 
We devise a new algorithm to solve this type of non-convex problems by formulating a rapidly-exploring random tree of barrier pairs, with each barrier pair composed of a quadratic barrier function and a full state feedback controller. 
The proposed method employs a rapid-exploring random tree to deal with the non-convex constraints and uses barrier pairs to fulfill the local convex constraints.
As such, the method solves control problems fulfilling the required transitions of an automaton in order to satisfy given linear temporal logic constraints. At the same time it synthesizes locally optimal controllers in order to transition between the regions corresponding to the alphabet of the automaton. 
We demonstrate this new algorithm on a simulation of a two linkage manipulator robot.
\end{abstract}

\section{Introduction}


Linear temporal logic ($\mathsf{LTL}$) helps control system designers to define specifications for controlling dynamical systems. 
Synthesizing a controller subject to an $\mathsf{LTL}$ specification usually starts with constructing a finite discrete abstraction of a dynamical system through a partition of the continuous state space. 
The atomic propositions ($\mathsf{AP}$s) associated with the temporal logic specification represent different regions of the partitioned state space.
Then, we can use formal synthesis methods to build a discrete controller for fulfilling the $\mathsf{LTL}$ specification.
However, a dynamical system with an $\mathsf{LTL}$ specification naturally leads to a hybrid control problem \cite{liu2014abstraction}.
To complete the $\mathsf{LTL}$ synthesis process for a dynamical system, we also need to find the low-level controllers (in the continuous state space) for executing the transitions between the abstract states of the high-level discrete controller.

Hybrid control that bridges the $\mathsf{LTL}$ specification and continuous state-space dynamics is a challenging problem, especially for nonlinear dynamical systems such as robots. 
By synthesizing barrier certificates through sum-of-squares optimization \cite{prajna2006barrier}, temporal logic specifications can be effectively verified when applied to nonlinear dynamical systems \cite{wongpiromsarn2015automata}. 
However, this work is not focused on control synthesis. 
In \cite{papusha2016automata}, an optimal control synthesis algorithm using approximate dynamic programming combines dynamical system variables and automata transitions into a single cost function. 
This method is used to synthesize continuous state trajectories that follow a deterministic finite automaton transferred from a co-safe $\mathsf{LTL}$ specification. 
Nevertheless, this work only addresses problems with convex state-space constraints.

The low-level control synthesis needs to guarantee that the transitions for all continuous states in one $\mathsf{AP}$ region to another $\mathsf{AP}$ region following the high-level discrete controller. It can be considered as a trajectory planning problem with uncertain initial state conditions corresponding to the regions defined by the atomic propositions. 
The region of attraction of the generated robust trajectory planner is also known as a `funnel'  \cite{burridge1999sequential}. 
A `funnel' can be synthesized over a shooting trajectory via multiple local stabilizing controllers \cite{tedrake2010lqr} or by solving quadratic programs based on control barrier functions \cite{ames2016control}. 
These strategies have been proposed to solve closed system problems \cite{nilsson2018barrier} and reactive synthesis problems \cite{decastro2015synthesis} with temporal logic constraints.
The real challenge is that the trajectory planning problem in its general form is a non-convex problem, for instance, when there are $\mathsf{AP}$ regions located between the initial and goal $\mathsf{AP}$ regions. 
In \cite{reist2016feedback}, a simulation-based method to solve the non-convex problem is proposed by simulating a number of `funnels' and checking constraint satisfaction for each funnel. 
However, simulation-based methods suffer from high computational costs.

\begin{figure}
    \centering
	\def\svgwidth{.48\textwidth}
	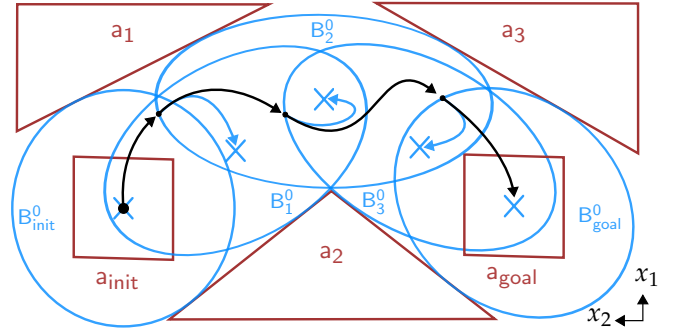
    \caption{Atomic proposition $\mathsf{a_{init}}$, $\mathsf{a_{1}}$, $\mathsf{a_{2}}$, $\mathsf{a_{3}}$, $\mathsf{a_{goal}}$ represent the polytopic regions (red) in workspace. The proposed $\mathsf{BP}$-$\mathsf{RRT}$ consists of a sequence of barrier pairs with their sub-level sets $\mathsf{B ^ {\leq 0}}$ (blue) interconnected to guarantee the transition between $\mathsf{a_{init}}$ and $\mathsf{a_{goal}}$. An example trajectory (black) switches to a different barrier pair controller as it enters the sub-level set of another barrier pair.}
    \label{figure}
\end{figure}

For robotic systems, this type of non-convex motion planning problem is usually addressed using sampling-based methods such as the rapidly-exploring random tree ($\mathsf{RRT}$) method \cite{lavalle2000rapidly}.
In an $\mathsf{RRT}$ algorithm, a random position $\mathrm{x}_\mathsf{rand}$ is sampled from the reachable space in every iteration. 
An $\mathsf{RRT}$ graph expands toward the sampled position from its closest vertex $\mathrm{x}_{\mathsf{near}}$ by a predefined distance $\delta$. 
The trajectory that connects an initial position $\mathrm{x}_{\mathsf{init}}$ and a desired position $\mathrm{x}_{\mathsf{goal}}$ can be extracted from the graph in the end.
By incorporating optimal control theory in the sampled trajectory of $\mathsf{RRT}$, the convergence rate of the motion planning problem is greatly improved \cite{karaman2011sampling}.
To improve the exploration of the $\mathsf{RRT}$ graph, a sampling strategy based on the estimated feasibility set of a robot is proposed in \cite{shkolnik2009reachability}.
However, the transitions between the vertices of the $\mathsf{RRT}$ trajectory are not guaranteed to avoid collisions with the undesirable state-space regions without having barrier certificates \cite{prajna2006barrier} along the $\mathsf{RRT}$ trajectory.

In order to solve the non-convex robot motion planning problem with $\mathsf{LTL}$ constraints, we propose a new approach consisting of a rapidly-exploring random tree of barrier pairs \cite{thomas2018safety}, where each barrier pair is composed of a quadratic barrier function and a full state feedback controller. Our method employs $\mathsf{RRT}$ to deal with non-convex constraints while employing barrier pairs equipped with sub-optimal stabilizing controllers to fulfill local convex constraints. By using our approach, a sequence of barrier pairs is effectively synthesized to execute the required transitions of an automaton that satisfy given $\mathsf{LTL}$ specifications. 
For validation, we implement our new approach on a simulation of a two-link manipulator robot.

\section{Preliminaries}

In this section, we recall the basics of multi-body robot dynamics, barrier pairs, rapidly-exploring random trees and nondeterministic Buchi automaton. For convenience, $\mathsf{a}_{\mathsf{i}}$ is defined as an atomic proposition corresponding to a region in the workspace of a robot, and $\bar{\mathsf{a}}_{\mathsf{i}} \triangleq \R ^ {\mathsf{n}} \smallsetminus \mathsf{a}_{\mathsf{i}}$ is defined as a workspace region excluding the set for $\mathsf{a}_{\mathsf{i}}$.

\subsection{Multi-Body Robot Dynamics}

The Lagrangian dynamics of an n-DOF robot can be expressed as
\begin{equation} \label{eq:lg}
\mathrm{M}(\mathrm{q}) \ddot{\mathrm{q}} + \mathrm{C}(\mathrm{q}, \, \dot{\mathrm{q}}) \dot{\mathrm{q}} = \mathrm{u}
\end{equation}
where $\mathrm{M}(\mathrm{q})$ is the matrix of inertia, $\mathrm{C}(\mathrm{q}, \, \dot{\mathrm{q}})$ is the coefficient matrix of Coriolis and centrifugal effects, $\mathrm{q} \triangleq [ q_1, \, \cdots, \, q_n ] ^ \top$ is the vector of joint positions with $\dot{\mathrm{q}}$ and $\ddot{\mathrm{q}}$ defined as its first and second order time derivatives and $\mathrm{u} \triangleq [ u_1, \, \cdots, \, u_n ] ^ \top$ is the vector of joint torques.
The n-dimensional workspace position vector $\mathrm{x} \triangleq [ x_1, \, \cdots, \, x_n ] ^ \top$ can be calculated from the joint position vector using
\begin{equation} \label{eq:forward}
\mathrm{x} = \mathrm{F(q)}
\end{equation}
where $\mathrm{F(\cdot)}$ represents the forward kinematics.
By linearizing \eqref{eq:lg} and \eqref{eq:forward} around an equilibrium point $[ \mathrm{q_e} ^ \top, \, \vec{0} ^ {\, \top}] ^ \top$, we obtain the state-space form
\begin{align}
\begin{bmatrix}
\dot{\tilde{\mathrm{q}}} \\
\ddot{\tilde{\mathrm{q}}}
\end{bmatrix}
& =
\begin{bmatrix}
\mathbf{0} & \mathbf{I} \\
\mathbf{0} & \mathrm{M}^{-1}(\mathrm{q}_\mathrm{e}) \mathrm{C}(\mathrm{q}_\mathrm{e}, \, \vec{0} )
\end{bmatrix}
\begin{bmatrix}
\tilde{\mathrm{q}} \\
\dot{\tilde{\mathrm{q}}}
\end{bmatrix}
+
\begin{bmatrix}
\mathbf{0} \\
\mathrm{M}^{-1}(\mathrm{q}_\mathrm{e})
\end{bmatrix}
\mathrm{u} \label{eq:ss-lg} \\
\tilde{\mathrm{x}} \,
& =
\begin{bmatrix}
\mathrm{J(q_e)} & \mathbf{0}
\end{bmatrix}
\begin{bmatrix}
\tilde{\mathrm{q}} \\
\dot{\tilde{\mathrm{q}}}
\end{bmatrix} \label{eq:ss-jacobian}
\end{align}
where $\tilde{\mathrm{q}} \triangleq \mathrm{q} - \mathrm{q_e}$ and $\tilde{\mathrm{x}} \triangleq \mathrm{x} - \mathrm{x_e}$ with $\mathrm{x_e} = \mathrm{F(q_e)}$. The partial derivative of $\mathrm{F(q)}$ with respect to $\mathrm{q}$ is the Jacobian matrix $\mathrm{J(q)}$.

\subsection{Barrier Pairs}

{\bf Definition 1} \cite{thomas2018safety}:
A \emph{barrier pair} is a pair consisting of a barrier function and a controller $(B,\ k)$ with the following properties
\begin{itemize}
\item[(a)] $-1<B(\mathrm{\tilde{q}}, \, \mathrm{\dot{\tilde{q}}})\leq 0, \mathrm u=k(\mathrm{\tilde{q}}, \, \mathrm{\dot{\tilde{q}}}) \implies \dot B(\mathrm{\tilde{q}}, \, \mathrm{\dot{\tilde{q}}}) < 0$,
\vspace{3pt}
\item[(b)] $B(\mathrm{\tilde{q}}, \, \mathrm{\dot{\tilde{q}}})\leq 0 \implies \mathrm{[ \tilde{q} ^ \top, \, \dot{\tilde{q}} ^ \top] ^ \top} \in \mathsf{Z},\ k(\mathrm{\tilde{q}}, \, \mathrm{\dot{\tilde{q}}}) \in \mathsf{U}$,
\vspace{1pt}
\end{itemize}
where $\mathrm{[ \tilde{q} ^ \top, \, \dot{\tilde{q}} ^ \top] ^ \top} \in \mathsf{Z}$ and $\mathrm{u} \in \mathsf{U}$ are the state and input constraints. These properties are also called the invariance and constraint satisfaction properties of a barrier pair. If we define the barrier pair as 
\begin{equation} \label{eq:bp}
B = 
\begin{bmatrix}
\tilde{\mathrm{q}} \\
\dot{\tilde{\mathrm{q}}}
\end{bmatrix} 
^ \top
\mkern-14mu
\mathrm{Q}^{-1} 
\mkern-6mu
\begin{bmatrix}
\tilde{\mathrm{q}} \\
\dot{\tilde{\mathrm{q}}}
\end{bmatrix}
- 1, \quad k = \mathrm{K} \begin{bmatrix}
\tilde{\mathrm{q}} \\
\dot{\tilde{\mathrm{q}}}
\end{bmatrix}
\end{equation}
where $B$ is a quadratic barrier function with a positive definite matrix $\mathrm{Q}$ and $k$ is a full state feedback controller,
the barrier pair synthesis becomes a linear matrix inequality ($\mathsf{LMI}$) optimization problem \cite{thomas2018safety}.
We define $\mathsf{B}^{\epsilon} \triangleq \{ \mathrm{[ \tilde{q} ^ \top, \, \dot{\tilde{q}} ^ \top] ^ \top} \mid B = \epsilon \}$ as the level set of $B$ corresponding to a value $\epsilon$ 
and $\mathsf{B}^{\leq \epsilon} \triangleq \{ \mathrm{[ \tilde{q} ^ \top, \, \dot{\tilde{q}} ^ \top] ^ \top} \mid B \leq \epsilon \}$ as the sub-level set of $B$ corresponding to $\epsilon$.

\begin{algorithm}[t] 
\caption{$G \leftarrow \texttt{RRT}(\mathrm{x}_{\mathsf{init}}, \, \mathrm{x}_{\mathsf{goal}}, \, \delta, \, \bar{\mathsf{a}}_{\mathsf{1}}, \, \cdots, \, \bar{\mathsf{a}}_{\mathsf{n_o}})$} \label{code:rrt}
\begin{algorithmic} [1]
\REQUIRE Initial state $\mathrm{x}_{\mathsf{init}}$, goal state $\mathrm{x}_{\mathsf{goal}}$, incremental distance $\delta$, state constraints $\bar{\mathsf{a}}_{\mathsf{1}}, \, \cdots, \, \bar{\mathsf{a}}_{\mathsf{n_o}}$
\ENSURE $\mathsf{RRT}$ graph $G$
\STATE $\delta_{\mathsf{0}} \leftarrow \texttt{GetDistance}(\mathrm{x}_{\mathsf{goal}}, \, \mathrm{x}_{\mathsf{init}})$
\STATE $G.\texttt{AddVertex}(\mathrm{x}_{\mathsf{goal}})$
\STATE $\mathrm{x}_{\mathsf{new}} \leftarrow \mathrm{x}_{\mathsf{goal}}$
\WHILE{$\delta_{\mathsf{0}} > \delta$}
\STATE $\mathrm{x}_{\mathsf{rand}} \leftarrow \texttt{RandomState}(\bigcap_{\mathsf{i=1}}^{\mathsf{n_o}} \bar{\mathsf{a}}_{\mathsf{i}})$
\STATE $\mathrm{x}_{\mathsf{near}} \leftarrow \texttt{NearestVertex}(\mathrm{x}_{\mathsf{rand}}, \, G)$
\STATE $\mathrm{x}_{\mathsf{new}} \leftarrow \texttt{NewState}(\mathrm{x}_{\mathsf{near}}, \, \mathrm{x}_{\mathsf{rand}}, \, \delta)$
\IF{$\mathrm{x}_{\mathsf{new}} \in \bigcap_{\mathsf{i=1}}^{\mathsf{n_o}} \bar{\mathsf{a}}_{\mathsf{i}}$}
\STATE $\delta_{\mathsf{0}} \leftarrow \texttt{GetDistance}(\mathrm{x}_{\mathsf{new}}, \, \mathrm{x}_{\mathsf{init}})$
\STATE $G.\texttt{AddVertex}(\mathrm{x}_{\mathsf{new}}), G.\texttt{AddEdge}((\mathrm{x}_{\mathsf{near}}, \mathrm{x}_{\mathsf{new}}))$
\ENDIF
\ENDWHILE
\STATE $G.\texttt{AddVertex}(\mathrm{x}_{\mathsf{init}}), G.\texttt{AddEdge}((\mathrm{x}_{\mathsf{new}}, \mathrm{x}_{\mathsf{init}}))$
\end{algorithmic}
\end{algorithm}

\subsection{Rapidly-Exploring Random Trees}

Let us recall the algorithm of $\mathsf{RRT}$ that generates trajectories from $\mathrm{x}_{\mathsf{init}}$ to $\mathrm{x}_{\mathsf{goal}}$ subject to workspace constraints $\mathrm{x} \in \bigcap_{\mathsf{i=1}}^{\mathsf{n_o}} \bar{\mathsf{a}}_{\mathsf{i}}$ where $\mathsf{n_o}$ is the number of undesirable regions. In Algorithm~\ref{code:rrt}, a random state $\mathrm{x}_\mathsf{rand}$ is sampled from the reachable space in line 5. In line 6-7, the graph extends toward the sampled state from its closest vertex by a constant distance $\delta$. The algorithm terminates when distance to the initial state $\mathrm{x}_{\mathsf{init}}$ is smaller than $\delta$. The trajectory that connects $\mathrm{x}_{\mathsf{init}}$ and $\mathrm{x}_{\mathsf{goal}}$ can be generated from the graph.

\subsection{Nondeterministic Buchi Automaton}

{\bf Definition 2}:  
A \emph{Nondeterministic Buchi automaton} $\mathsf{A = (S, \, 2 ^ {AP}, \, d, \, S_0, \, S_f)}$ is a tuple where 
\begin{itemize}
\item[(a)] $\mathsf{S}$ is a set of discrete states,
\item[(b)] $\mathsf{2 ^ {AP}}$ is the power set of atomic propositions,
\item[(c)] $\mathsf{d: S \times 2 ^ {AP}} \rightarrow  \mathsf{2 ^ S}$ is a transition function,
\item[(d)] $\mathsf{S_0 \subseteq S}$ is a set of initial states, and
\item[(e)] $\mathsf{S_f \subseteq S}$ is a set of accept states.
\end{itemize}
A $\mathsf{LTL}$ specification $\varphi$ can be transformed into a nondeterministic Buchi automaton and satisfied by an accepting run (of transitions) of its corresponding nondeterministic Buchi automaton if the run visits a state in $\mathsf{S_f}$ infinitely often.

\section{Problem Statement}

In this paper, we consider a robot with its workspace trajectories constrained by a $\mathsf{LTL}$ specification. The $\mathsf{LTL}$ specification is defined based on $\mathsf{AP}$s that represent different polytopic regions in the workspace of the robot. 

{\bf Problem}:
For a given linear temporal logic specification $\phi$, 
find a rapidly-exploring random tree of barrier pairs such that the robot fulfills an accepting run of a nondeterministic Buchi automaton that represents $\phi$.

\section{Methods}


The proposed approach is illustrated in Fig.~\ref{figure}. 
It starts with finding an equilibrium point inside the goal $\mathsf{AP}$ region $\mathsf{a}_{\mathsf{goal}}$ and synthesizing a barrier pair in the form of \eqref{eq:bp}, subject to local convex state constraints (e.g. surrounding undesirable $\mathsf{AP}$ regions) for this equilibrium. 
We then sample a new equilibrium point inside the sub-level set $\mathsf{B_{goal}}^{\leq 0}$ for the first barrier pair and synthesize a new barrier pair subject again to local convex state constraints for the new equilibrium. 
Inside the sub-level sets of the existing barrier pairs, another equilibrium is sampled, followed by a barrier pair synthesis. 
This barrier pair sampling process is iterated until the sub-level set of a barrier pair contains the equilibrium of a barrier pair whose sub-level set $\mathsf{B_{init}}^{\leq 0}$ contains the entire initial $\mathsf{AP}$ region $\mathsf{a_{init}}$. 
In the end, we obtain a sequence of interconnected barrier pairs between $\mathsf{a}_{\mathsf{init}}$ and $\mathsf{a}_{\mathsf{goal}}$ without passing through undesirable $\mathsf{AP}$ regions.

\subsection{Norm-Bound Linear Differential Inclusion Model}

Our proposed method relies on formulating an $\mathsf{LMI}$ problem to synthesize the barrier pairs subject to local convex constraints. However, the linearized state space equations \eqref{eq:ss-lg} and \eqref{eq:ss-jacobian} become inaccurate if the state $[ \mathrm{q} ^ \top, \, \dot{\mathrm{q}} ^ \top ] ^ \top$ deviates from the equilibrium. Before employing barrier pair synthesis, we need to ensure that the linear model is valid for all states in the constrained state space $\mathsf{Z}$ of the barrier pair. 

If we express the norm-bound uncertainties of the linearized robot dynamical model in \eqref{eq:ss-lg} and \eqref{eq:ss-jacobian} as 
\begin{align}
\mathrm{M} ^ {-1} (\mathrm{q}) \mathrm{C}(\mathrm{q}, \, \dot{\mathrm{q}}) & \in \{ \mathrm{A_1} + \mathrm{A_2} \Delta \mathrm{A_3}: \ \norm{\Delta} \leq 1 \} \label{eq:As} \\
\mathrm{M} ^ {-1} (\mathrm{q}) & \in \{ \mathrm{B_1} + \mathrm{B_2} \Delta \mathrm{B_3}: \ \norm{\Delta} \leq 1 \} \label{eq:Bs} \\
\mathrm{J} (\mathrm{q}) & \in \{ \mathrm{J_1} + \mathrm{J_2} \Delta \mathrm{J_3}: \ \norm{\Delta} \leq 1 \} \label{eq:Js} 
\end{align}
for all state $[ \mathrm{q} ^ \top, \, \dot{\mathrm{q}} ^ \top ] ^ \top$ in the constrained state space $\mathsf{Z}$ around the equilibrium,
a norm-bound linear differential inclusion (LDI) \cite{boyd1994linear} that is valid for all states in $\mathsf{Z}$ can be expressed as
\begin{align}
\begin{bmatrix}
\dot{\tilde{\mathrm{q}}} \\
\ddot{\tilde{\mathrm{q}}}
\end{bmatrix}
& =
\begin{bmatrix}
\mathbf{0} & \mathbf{I} \\
\mathbf{0} & \mathrm{A_1} + \mathrm{A_2} \Delta \mathrm{A_3}
\end{bmatrix}
\begin{bmatrix}
\tilde{\mathrm{q}} \\
\dot{\tilde{\mathrm{q}}}
\end{bmatrix}
+
\begin{bmatrix}
\mathbf{0} \\
\mathrm{B_1} + \mathrm{B_2} \Delta \mathrm{B_3}
\end{bmatrix}
\mathrm{u} \label{eq:ss-lg-robust} \\
\tilde{\mathrm{x}} \,
& =
\begin{bmatrix}
\mathrm{J_1} + \mathrm{J_2} \Delta \mathrm{J_3} & \mathbf{0}
\end{bmatrix}
\begin{bmatrix}
\tilde{\mathrm{q}} \\
\dot{\tilde{\mathrm{q}}}
\end{bmatrix}. \label{eq:ss-jacobian-robust}
\end{align}
One way of finding the norm-bound LDI is to calculate $\mathrm{M} ^ {-1} (\mathrm{q}) \mathrm{C}(\mathrm{q}, \, \dot{\mathrm{q}})$, $\mathrm{M} ^ {-1} (\mathrm{q})$ and $\mathrm{J} (\mathrm{q})$ from a number of sample states in $\mathsf{Z}$ and use quadric inclusion programs \cite{thomas2019quadric} to fit an inclusion model.


Since we assume the $\mathsf{AP}$ regions are polytopic, each edge of an $\mathsf{AP}$ region can be transformed into an inequality constraint. To exclude the undesirable regions of a transition, only one of these inequality constraints need to be considered for each undesirable region. Otherwise, the state space can be over-constrained. If the workspace position $\mathrm{x_e}$ of an equilibrium satisfies multiple inequality constraints associated with an undesirable region, we can select the edge which has the maximal distance to $\mathrm{x_e}$ to avoid being over-constrained. 
Based on the selected inequality constraints $|\mathrm{a_i}  \tilde{\mathrm{x}}| < \bar{a}_\mathrm{i}$ associated with all undesirable regions $\mathsf{a_1, \, a_2, \, \cdots, \, a_{n_o}}$, a local convex state space region $\mathsf{Z_{safe}}$ can be defined as
\begin{equation} \label{eq:Qsafe}
\begin{aligned}
\mathsf{Z_{safe}} 
\triangleq 
\{ 
\mathrm{[ \tilde{q} ^ \top, \, \dot{\tilde{q}} ^ \top] ^ \top} 
:
\abs{\mathrm{a_i}  (\mathrm{J_1} + \mathrm{J_2} \Delta \mathrm{J_3}) \ \mathrm{\tilde{q}}} < \bar{a}_\mathrm{i}, \\
\norm{\Delta} \leq 1, \
\mathrm{i = 1, \, \cdots, \, n_o}
\},
\end{aligned}
\end{equation}
where $\mathrm{a_i}$ for $\mathrm{i = 1, \, \cdots, \, n_o}$ are row vectors with $\mathrm{n_o}$ as the number of undesirable $\mathsf{AP}$ regions.

\begin{figure}
    \centering
	\def\svgwidth{.48\textwidth}
	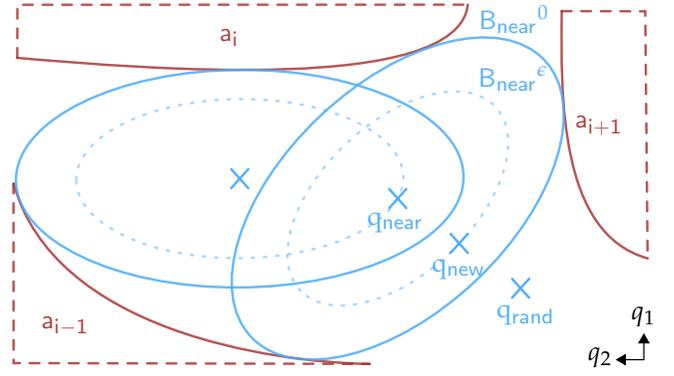
    \caption{By projecting a random joint space position $\mathrm{q}_{\mathsf{rand}}$ to the hyper-surface of $\mathsf{B}_{\mathsf{near}}^{\epsilon}$ of the nearest barrier pair, a new equilibrium of $\mathsf{BP}$-$\mathsf{RRT}$ is created. Notice that even if the workspace undesirable regions are polytopic, their joint space projections are not guaranteed to be also polytopic.}
    \label{fig:new-equilibrium}
\end{figure}

However, $\mathsf{Z_{safe}}$ cannot be directly used as the constrained state space region $\mathsf{Z}$ for barrier pair synthesis because it has no joint velocity state constraints. In some cases, a selected edge of an undesirable region has a very long distance to $\mathrm{x_e}$ and result in a very large uncertainty of the norm-bound LDI model. Therefore, we need some additional state space constraints for defining $\mathsf{Z}$. 
Let us first define another constrained state space $\mathsf{Z_0}$ as
\begin{equation} \label{eq:Q0}
\begin{aligned}
\mathsf{Z_0} 
\triangleq 
\{ 
\mathrm{[ \tilde{q} ^ \top, \, \dot{\tilde{q}} ^ \top] ^ \top} 
:
\abs{\mathrm{b_i}  (\mathrm{J_1} + \mathrm{J_2} \Delta \mathrm{J_3}) \ \mathrm{\tilde{q}}} < \bar{x}_\mathrm{i}, \
\abs{\mathrm{b_i}  \mathrm{\dot{\tilde{q}}}} < \bar{\dot q}_\mathrm{i}, \\
\norm{\Delta} \leq 1, \
\mathrm{i = 1, \, \cdots, \, n}
\},
\end{aligned}
\end{equation}
where $\mathrm{b_i}$ for $\mathrm{i = 1, \, \cdots, \, n}$ are the standard basis (row) vectors of $\mathrm{n}$-dimensional Euclidean space.
Then, the constrained state space region for the valid norm-bound LDI model is defined as $\mathsf{Z} \triangleq \mathsf{Z_{safe}} \cap \mathsf{Z_0}$.

Similar to \eqref{eq:Qsafe} and \eqref{eq:Q0}, the constrained input space region $\mathsf{U}$ for the barrier pair synthesis can be defined as
\begin{equation} \label{eq:U}
\begin{aligned}
\mathsf{U} 
\triangleq 
\{ 
\mathrm{u} 
:
\abs{\mathrm{b_i}  \mathrm{u}} < \bar{u}_\mathrm{i}, \
\mathrm{i = 1, \, \cdots, \, n}
\}.
\end{aligned}
\end{equation}

\subsection{Barrier Pair Synthesis Sub-Problems}

The barrier pair synthesis problem includes a series of $\mathsf{LMI}$ constraints and generates a quadratic barrier function $B$ with a full state controller $k$ in the form of \eqref{eq:bp}. 
First, the sequence of barrier pairs needs to contain the two desired $\mathsf{AP}$ regions $\mathsf{a_{init}}$ and $\mathsf{a_{goal}}$ defining the automaton transitions and exclude all undesirable $\mathsf{AP}$ regions $\mathsf{a_1, \, \cdots, \, a_{n_o}}$. 

Although a desired $\mathsf{AP}$ region $\mathsf{a_{d}}$ is assumed to be polytopic in the Cartesian workspace, its joint space projection is not guaranteed to be polytopic. In order to ensure that the ellipsoidal sub-level set $\mathsf{B}^{\leq 0}$ of a barrier pair contains $\mathsf{a_{d}}$, we sample a number of points from all edges of $\mathsf{a_{d}}$ and let $\mathsf{B}^{\leq 0}$ contain the joint space projections of these Cartesian space samples using the following set of $\mathsf{LMI}$s
\begin{equation} \label{eq:x-inclusion}
\begin{bmatrix}
1 & \star \\
 \mathrm{R(x_i)} - \mathrm{q_e} & \mathrm{S_1} \mathrm{Q}  \mathrm{S_1} ^ \top 
\end{bmatrix} \succeq 0, \quad \forall \ \mathrm{i = 1, \dotsc, n_p} 
\end{equation}
where $\mathrm{n_p}$ is the number of sampled workspace points at the edge of $\mathsf{a}_{\mathsf{d}}$ with $\mathsf{a}_{\mathsf{d}} = \mathsf{Co} \{\mathrm{x_1}, \, \cdots, \, \mathrm{x_p} \}$, $\mathrm{R(\cdot)}$ is an inverse kinematics operator and $\mathrm{S_1} \triangleq \mathrm{[ I_{n \times n}, \, 0_{n \times n}]}$.

Using the $\mathsf{S}$-procedure \cite{ma2006lmi}, the inequality constraints $|\mathrm{a_i}  \tilde{\mathrm{x}}| < \bar{a}_\mathrm{i}$ of $\mathsf{Z_{safe}}$ in \eqref{eq:Qsafe} can be transformed into a set of $\mathsf{LMI}$s
\begin{equation} \label{eq:x-exclusion}
\begin{aligned}
\begin{bmatrix}
\bar{a}_\mathrm{i}^{2} \mathrm{Q} & \star & \star & \star \\
\mathbf{0} & \upgamma_\mathrm{i} \mathbf{I} & \star & \star \\
\mathrm{a_i J_1 S_1 Q} & \upgamma_\mathrm{i} \mathrm{a_i J_2} & 1 & \star \\
\mathrm{J_3 S_1 Q} & \mathbf{0} & \vec{0} & \upgamma_\mathrm{i} \mathbf{I}
\end{bmatrix} \succeq 0, \ \
\forall \  \mathrm{i = 1, \dotsc, n_o} 
\end{aligned}
\end{equation}
where $\upgamma_\mathrm{i}$ for $\mathrm{i = 1, \dotsc, n_o}$ are positive real scalar variables. 

Similar to \eqref{eq:x-exclusion}, the workspace position constraints $|\mathrm{b_i}  \tilde{\mathrm{x}}| < \bar{x}_\mathrm{i}$ of $\mathsf{Z_0}$ defined in \eqref{eq:Q0} can be transformed into a set of $\mathsf{LMI}$s 
\begin{equation} \label{eq:x-limit}
\begin{aligned}
\begin{bmatrix}
\bar{x}_\mathrm{i}^{2} \mathrm{Q} & \star & \star & \star \\
\mathbf{0} & \upmu_\mathrm{i} \mathbf{I} & \star & \star \\
\mathrm{b_i J_1 S_1 Q} & \upmu_\mathrm{i} \mathrm{b_i J_2} & 1 & \star \\
\mathrm{J_3 S_1 Q} & \mathbf{0} & \vec{0} & \upmu_\mathrm{i} \mathbf{I}
\end{bmatrix} \succeq 0, \ \
\forall \  \mathrm{i = 1, \dotsc, n} 
\end{aligned}
\end{equation}
where $\upmu_\mathrm{i}$ for $\mathrm{i = 1, \dotsc, n}$ are positive real scalar variables. The joint velocity constraint $\mathsf{LMI}$s of $\mathsf{Z_0}$ are expressed as
\begin{equation} \label{eq:qdot-limit}
\begin{bmatrix}
\mathrm{Q} & \star \\
\mathrm{b_i S_2 Q} & \bar{\dot{q}}_\mathrm{i} ^ 2
\end{bmatrix} \succeq 0, \quad \forall \  \mathrm{i = 1, \dotsc, n} 
\end{equation}
where $\mathrm{S_2} \triangleq \mathrm{[ 0_{n \times n}, \, I_{n \times n}]}$.

Although the full state feedback controller $k$ in \eqref{eq:bp} turns the input constraints into state constraints, $\mathrm{K}$ is also a variable to be solved. In \cite{boyd1994linear}, a new variable $\mathrm{Y} \triangleq \mathrm{KQ}$ is introduced to express the input constraints into $\mathsf{LMI}$s. After the barrier pair synthesis problem is solved, $\mathrm{K}$ can be extracted by multiplying $\mathrm{Y}$ by $\mathrm{Q} ^ {-1}$ on the right hand side.
The input constraint $\mathsf{LMI}$s can be expressed as
\begin{equation} \label{eq:u-limit}
\begin{bmatrix}
\mathrm{Q} & \star \\
\mathrm{b_i Y} & \bar{u}_\mathrm{i} ^ 2
\end{bmatrix} \succeq 0, \quad \forall\  \mathrm{i = 1, \dotsc, n}
\end{equation}
for enforcing the input constraints $|\mathrm{b_i u}|\leq \bar{u}_\mathrm{i}$ of $\mathsf{U}$ defined in \eqref{eq:U}.

To guarantee the invariance of the barrier function, we include a Lyapunov stability $\mathsf{LMI}$ in \cite{boyd1994linear} for the norm-bound LDI model
\begin{equation} \label{eq:stability}
\begin{aligned}
\begin{bmatrix}
\mathrm{H} + 2 \alpha \mathrm{Q} & \star & \star \\
\mathrm{A_3 S_2 Q} & - \upmu_\mathrm{x} \mathbf{I} & \star \\
\mathrm{B_3 Y} & \mathbf{0} & - \upmu_\mathrm{u} \mathbf{I} \\
\end{bmatrix} \preceq 0,
\end{aligned}
\end{equation}
where $\upmu_\mathrm{x}$ and $\upmu_\mathrm{u}$ are positive real scalar variables, $\alpha$ is a basic decay rate of the barrier function and $\mathrm{H}$ is defined as
\begin{equation} 
\begin{aligned}
\mathrm{H} 
\triangleq 
 \mathsf{He} \{ \mathrm{S_1 ^ \top S_2 Q + S_2 ^ \top A_1 S_2 Q + S_2 ^ \top B_1 Y} \} \\ 
 + \upmu_\mathrm{x} \mathrm{S_2 ^ \top A_2 A_2 ^ \top S_2} 
+ \upmu_\mathrm{u} \mathrm{S_2 ^ \top B_2 B_2 ^ \top S_2}
\end{aligned}
\end{equation}
where $\mathsf{He} \{ \star \} \triangleq \star + \star ^ \top$.

\begin{algorithm}[t]
\caption{$G \leftarrow \texttt{BPRRT}(\mathsf{a}_{\mathsf{init}}, \mathsf{a}_{\mathsf{goal}}, \epsilon, \bar{\mathsf{a}}_{\mathsf{1}}, \cdots, \bar{\mathsf{a}}_{\mathsf{n_o}}, \mathsf{Z_0}, \mathsf{U})$} \label{code:BP-RRT}
\begin{algorithmic} [1]
\REQUIRE Initial $\mathsf{AP}$ region $\mathsf{a}_{\mathsf{init}}$, goal $\mathsf{AP}$ region $\mathsf{a}_{\mathsf{goal}}$, barrier function threshold $\epsilon$, constraints associated with undesirable $\mathsf{AP}$ regions $\bar{\mathsf{a}}_{\mathsf{1}}, \, \cdots, \, \bar{\mathsf{a}}_{\mathsf{n_o}}$, state space constraint $\mathsf{Z_0}$, input constraint $\mathsf{U}$
\ENSURE $\mathsf{BP}$-$\mathsf{RRT}$ graph $G$
\STATE $\mathrm{x}_{\mathsf{init}} \leftarrow \texttt{GeometricCenter}(\mathsf{a}_{\mathsf{init}})$
\STATE $(B_{\mathsf{init}}, \, k_{\mathsf{init}}) \leftarrow \texttt{BP} (\mathrm{x}_{\mathsf{init}}, \, \mathsf{a}_{\mathsf{init}}, \, \bar{\mathsf{a}}_{\mathsf{1}}, \, \cdots, \, \bar{\mathsf{a}}_{\mathsf{n_o}}, \, \mathsf{Z_0}, \, \mathsf{U})$
%
\STATE $\mathrm{x}_{\mathsf{goal}} \leftarrow \texttt{GeometricCenter}(\mathsf{a}_{\mathsf{goal}})$
\STATE $(B_{\mathsf{goal}}, \, k_{\mathsf{goal}}) \leftarrow \texttt{BP} (\mathrm{x}_{\mathsf{goal}}, \, \mathsf{a}_{\mathsf{goal}}, \, \bar{\mathsf{a}}_{\mathsf{1}}, \, \cdots, \, \bar{\mathsf{a}}_{\mathsf{n_o}}, \, \mathsf{Z_0}, \, \mathsf{U})$
%
\STATE $G.\texttt{AddVertex}(\mathrm{x}_{\mathsf{goal}}), \, G.\texttt{AddBP}((B_{\mathsf{goal}}, \, k_{\mathsf{goal}}))$
\STATE $(B_{\mathsf{new}}, \, k_{\mathsf{new}}) \leftarrow (B_{\mathsf{goal}}, \, k_{\mathsf{goal}})$, $\mathrm{x}_{\mathsf{new}} \leftarrow \mathrm{x}_{\mathsf{goal}}$
\WHILE{$\mathrm{x}_{\mathsf{init}} \notin \mathsf{B}_{\mathsf{new}}^{\leq \epsilon}$}
\STATE $\mathrm{q}_{\mathsf{rand}} \leftarrow \texttt{RandomJointSpacePosition}(\R ^ {\mathrm n})$
\STATE $\mathrm{x}_{\mathsf{rand}} \leftarrow \texttt{ForwardKinematics}(\mathrm{q}_{\mathsf{rand}})$
\IF{$\mathrm{x}_{\mathsf{rand}} \in \bigcap_{\mathsf{i=1}}^{\mathsf{n_o}} \bar{\mathsf{a}}_{\mathsf{i}}$}
\STATE $\mathrm{q}_{\mathsf{near}}, \, \mathsf{B}_{\mathsf{near}} ^ {\epsilon} \leftarrow \texttt{NearestBP}(\mathrm{q}_{\mathsf{rand}}, \, G, \, \epsilon)$
\STATE $\mathrm{q}_{\mathsf{new}} \leftarrow \texttt{NewEquilibrium}(\mathrm{q}_{\mathsf{near}}, \, \mathrm{q}_{\mathsf{rand}}, \, \mathsf{B}_{\mathsf{near}} ^ {\epsilon})$
\STATE $\mathrm{x}_{\mathsf{new}} \leftarrow \texttt{ForwardKinematics}(\mathrm{q}_{\mathsf{new}})$
\STATE $(B_{\mathsf{new}}, \, k_{\mathsf{new}}) \leftarrow \texttt{BP} (\mathrm{x}_{\mathsf{new}}, \emptyset, \bar{\mathsf{a}}_{\mathsf{1}}, \cdots, \bar{\mathsf{a}}_{\mathsf{n_o}}, \mathsf{Z_0}, \mathsf{U})$
\STATE $G.\texttt{AddVertex}(\mathrm{x}_{\mathsf{new}}), \, G.\texttt{AddBP}((B_{\mathsf{new}}, \, k_{\mathsf{new}})),$
$G.\texttt{AddEdge}((\mathrm{x}_{\mathsf{near}}, \, \mathrm{x}_{\mathsf{new}}))$
\ENDIF
\ENDWHILE
\STATE $G.\texttt{AddVertex}(\mathrm{x}_{\mathsf{init}}), \, G.\texttt{AddBP}((B_{\mathsf{init}}, \, k_{\mathsf{init}})),$ $G.\texttt{AddEdge}((\mathrm{x}_{\mathsf{new}}, \, \mathrm{x}_{\mathsf{init}}))$
\end{algorithmic}
\end{algorithm}

Finally, the volume of the ellipsoid $\mathsf{B} ^ {\leq 0}$ is maximized through the cost function of the log of the determinant of $\mathrm{Q}$ \cite{boyd1994linear}. 
A barrier pair synthesis sub-problem $(B, \, k) = \texttt{BP} (\mathrm{x_e}, \, \mathsf{a}_{\mathsf{d}}, \, \bar{\mathsf{a}}_{\mathsf{1}}, \, \cdots, \, \bar{\mathsf{a}}_{\mathsf{n_o}}, \, \mathsf{Z_0}, \, \mathsf{U})$ for finding a sub-level set $\mathsf{B} ^ {\leq 0}$ that contains the desired $\mathsf{AP}$ region $\mathsf{a}_{\mathsf{d}}$ and excludes the undesirable $\mathsf{AP}$ regions $\mathsf{a_1, \, a_2, \, \cdots, \, a_{n_o}}$ can be expressed as
\begin{equation} \label{optimization 2}
\begin{aligned}
& \underset{\mathrm{Q, \, Y}}{\mathsf{maximize}}
& & \mathsf{log}(\mathsf{det}(\mathrm{Q})) \\
& \mathsf{subject \ to} & & \mathrm{Q} \succ 0, \\
&&& \eqref{eq:x-inclusion}, \, \eqref{eq:x-exclusion}, \, \eqref{eq:x-limit}, \, \eqref{eq:qdot-limit}, \, \eqref{eq:u-limit}, \, \eqref{eq:stability} \\
\end{aligned}
\end{equation}
which automatically generates a barrier pair $(B, k)$ if the problem is feasible.

\subsection{Barrier Pair Sampling Algorithm}

In Algorithm~\ref{code:rrt}, line 5-7 can be considered as the essential steps of building a $\mathsf{RRT}$ trajectory with the rest of the algorithm checking the state constraint satisfaction and the distance to $\mathrm{x}_{\mathsf{init}}$.
We leverage these essential steps of $\mathsf{RRT}$ to combine the barrier pair into a sequence that connects two $\mathsf{AP}$ regions in the reachable workspace. 

Algorithm~\ref{code:BP-RRT} describes our barrier pair rapidly-exploring random tree ($\mathsf{BP}$-$\mathsf{RRT}$) method. 
Line 1-6 in Algorithm~\ref{code:BP-RRT} initialize the graph by creating two barrier pairs which contain workspace regions $\mathsf{a}_{\mathsf{init}}$ and $\mathsf{a}_{\mathsf{goal}}$. 
The graph starts from the barrier pair of $\mathsf{a_{goal}}$.
In order to build the graph, a joint position $\mathrm{q}_\mathsf{rand}$ is sampled in line 8. If a sample of $\mathrm{q}_\mathsf{rand}$ is not reachable because of the undesirable $\mathsf{AP}$ regions, it will be excluded from the rest of the algorithm in line 10.

Line 11-14 in Algorithm~\ref{code:BP-RRT} is similar to line 5-7 in Algorithm~\ref{code:rrt}. 
However, instead of applying a constant incremental distance $\delta$ as $\mathsf{RRT}$, the new equilibrium $\mathrm{q}_{\mathsf{new}}$ is obtained by projecting the random equilibrium $\mathrm{q}_{\mathsf{rand}}$ to the hyper-surface of level set $\mathsf{B}_{\mathsf{near}} ^ {\epsilon}$ of the nearest barrier pair with $-1 < \epsilon \leq 0$ (Fig.~\ref{fig:new-equilibrium}). 
Therefore, $\mathrm{q}_{\mathsf{new}}$ is always inside the boundaries of the previously created barrier pairs and there is no need to check if $\mathrm{q}_{\mathsf{new}}$ satisfies the $\mathsf{AP}$ constraints. 

The algorithm terminates if there exists a sub-level set $\mathsf{B}_{\mathsf{new}}^{\leq \epsilon}$ of a new barrier pair that contains the equilibrium of the barrier pair of $\mathsf{a_{init}}$.
Then, the branch that connects $\mathsf{a}_{\mathsf{init}}$ and $\mathsf{a}_{\mathsf{goal}}$ can be extracted from the $\mathsf{BP}$-$\mathsf{RRT}$ graph. The barrier pair sequence is executed in reverse order for barrier pair synthesis to achieve the transition from $\mathsf{a_{init}}$ to $\mathsf{a_{goal}}$.



\section{Example}

\begin{figure}
    \small
    \centering
	\def\svgwidth{.45\textwidth}
	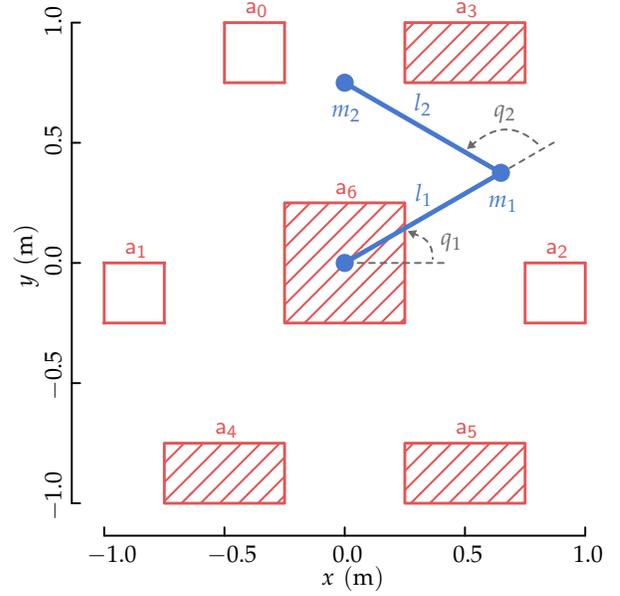
    \caption{A 2-link manipulator robot (blue) moves its end effector in a workspace with $\mathsf{AP}$ regions (red).}
    \label{fig:manipulator}
\end{figure}

Our $\mathsf{BP}$-$\mathsf{RRT}$ algorithm is demonstrated through a simulation of a 2-link manipulator robot with an equal length of $0.75 \, \mathrm{m}$ for each link, a mass of $2.5 \, \mathrm{kg}$ located at the distal end of each link, and a torque limit of $25 \, \mathrm{N \cdot m}$ for each joint. 
Fig.~\ref{fig:manipulator} shows the definition of $\mathsf{AP}$s in the workspace of the robot end effector, where $\mathsf{a_0}$, $\mathsf{a_1}$, $\mathsf{a_2}$ represent the desired task regions, $\mathsf{a_3}$, $\mathsf{a_4}$, $\mathsf{a_5}$ represent obstacle regions, and $\mathsf{a_6}$ represents the region where the robot's base is located.
A $\mathsf{LTL}$ specification $\phi$ is defined as
\begin{equation}
\begin{aligned}
& \phi \triangleq \phi_{\mathsf{init}} \wedge \phi_{\mathsf{live}} \wedge \phi_{\mathsf{safe}} \\
& \phi_{\mathsf{init}} \triangleq \mathsf{a_0} \\
& \phi_{\mathsf{live}} \triangleq \square \lozenge \mathsf{a_0} \wedge \square \lozenge \mathsf{a_1} \wedge \square \lozenge \mathsf{a_2} \\
& \phi_{\mathsf{safe}} \triangleq \square (\mathsf{free} \, \U \, (\mathsf{a_0} \vee \mathsf{a_1} \vee \mathsf{a_2})) \\
\end{aligned}
\end{equation}
where $\square$, $\lozenge$, and $\U$ are the $\mathsf{LTL}$ operators representing `always', `eventually', and `until' \cite{baier2008principles}. In addition, we define $\mathsf{free} \triangleq \neg \bigvee\limits_{\mathsf{i}=0}^6 \mathsf{a}_{\mathsf{i}}$.
Although the robot's end effector moves in a convex workspace region $\mathsf{X}_{\mathsf{convex}} \triangleq \{ [x, \, y] ^ \top: x ^ 2 + y ^ 2 \leq 2.25 \}$, the $\mathsf{free}$ workspace region $\mathsf{X_{free}} \triangleq \mathsf{X}_{\mathsf{convex}} \smallsetminus \bigcup\limits_{\mathsf{i=0}}^{\mathsf{6}} {\mathsf{a_i}}$ is non-convex. 

\begin{figure}[!tbp]
  	\centering
\begin{tikzpicture}[scale=0.16]
\tikzstyle{every node}+=[inner sep=0pt]
\draw [black] (27.9,-21.2) circle (3);
\draw (27.9,-21.2) node {\large $\mathsf{s_1}$};
\draw [black] (27.9,-32.8) circle (3);
\draw (27.9,-32.8) node {\large $\mathsf{s_4}$};
\draw [black] (47.1,-21.2) circle (3);
\draw (47.1,-21.2) node {\large $\mathsf{s_2}$};
\draw [black] (47.1,-32.8) circle (3);
\draw (47.1,-32.8) node {\large $\mathsf{s_3}$};
\draw [black] (47.1,-32.8) circle (2.4);
\draw [black] (15.8,-21.2) circle (3);
\draw (15.8,-21.2) node {\large $\mathsf{s_0}$};
\draw [black] (45.777,-18.52) arc (234:-54:2.25);
\draw (47.1,-13.95) node [above] {$\mathsf{a_0} \vee \mathsf{a_1} \vee \mathsf{free}$};
\fill [black] (48.42,-18.52) -- (49.3,-18.17) -- (48.49,-17.58);
\draw [black] (30.9,-21.2) -- (44.1,-21.2);
\fill [black] (44.1,-21.2) -- (43.3,-20.7) -- (43.3,-21.7);
\draw (37.5,-20.7) node [above] {$\mathsf{a_1}$};
\draw [black] (18.8,-21.2) -- (24.9,-21.2);
\fill [black] (24.9,-21.2) -- (24.1,-20.7) -- (24.1,-21.7);
\draw (21.85,-20.7) node [above] {$\mathsf{a_0}$};
\draw [black] (7.6,-21.2) -- (12.8,-21.2);
\draw (7.1,-21.2) node [left] {$\mathsf{start}$};
\fill [black] (12.8,-21.2) -- (12,-20.7) -- (12,-21.7);
\draw [black] (26.577,-18.52) arc (234:-54:2.25);
\draw (27.9,-13.95) node [above] {$\mathsf{a_0} \vee \mathsf{a_2} \vee \mathsf{free}$};
\fill [black] (29.22,-18.52) -- (30.1,-18.17) -- (29.29,-17.58);
\draw [black] (47.1,-24.2) -- (47.1,-29.8);
\fill [black] (47.1,-29.8) -- (47.6,-29) -- (46.6,-29);
\draw (47.6,-27) node [right] {$\mathsf{a_2}$};
\draw [black] (25.22,-34.123) arc (-36:-324:2.25);
\draw (20.65,-32.8) node [left] {$\mathsf{a_1} \vee \mathsf{a_2} \vee \mathsf{free}$};
\fill [black] (25.22,-31.48) -- (24.87,-30.6) -- (24.28,-31.41);
\draw [black] (44.1,-32.8) -- (30.9,-32.8);
\fill [black] (30.9,-32.8) -- (31.7,-33.3) -- (31.7,-32.3);
\draw (37.5,-33.3) node [below] {$\mathsf{a_1} \vee \mathsf{a_2} \vee \mathsf{free}$};
\draw [black] (27.9,-29.8) -- (27.9,-24.2);
\fill [black] (27.9,-24.2) -- (27.4,-25) -- (28.4,-25);
\draw (27.4,-27) node [left] {$\mathsf{a_0}$};
\end{tikzpicture}
    \caption{A Buchi automaton represents $\mathsf{LTL}$ specification $\phi$.}
    \label{fig:automata}
\end{figure}
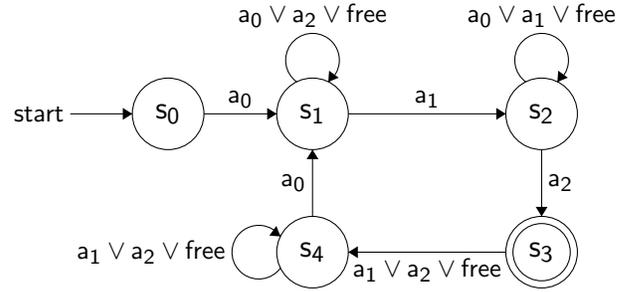

Fig.~\ref{fig:automata} shows a Buchi automaton corresponding to $\phi$.
An accepting run of the Buchi automaton starts from $\mathsf{a_0}$ and repeats the sequence of $\mathsf{a_0}, \, \mathsf{free}, \, \mathsf{a_1}, \, \mathsf{free}, \, \mathsf{a_2}, \, \mathsf{free}$ such that the accept state $\mathsf{s_3}$ of the Buchi automata is visited infinitely often. 

\begin{figure*}
    \footnotesize 
    \centering
	\def\svgwidth{1.\textwidth}
	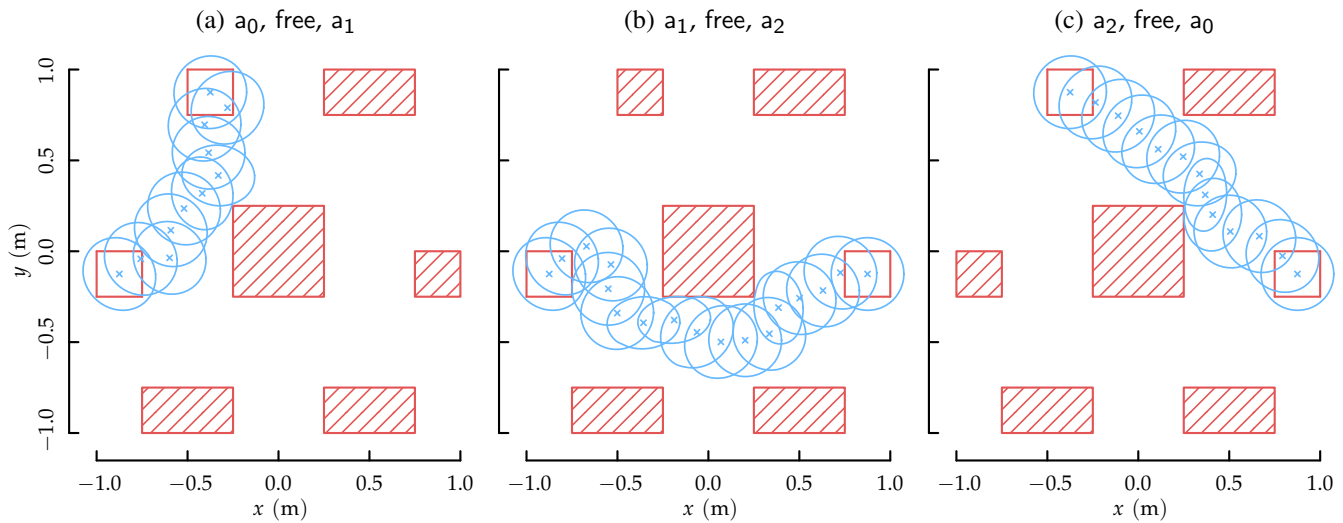
    \caption{Barrier pair sub-level sets (blue) connect $\mathsf{a_{init}}$ and $\mathsf{a_{goal}}$ (red non-striped) and avoid passing through the undesirable $\mathsf{AP}$ regions (red striped).}
    \label{fig:funnels}
\end{figure*}

In order to implement this accepting run of the Buchi automaton, we use $\mathsf{BP}$-$\mathsf{RRT}$ to build barrier pair sequences from $\mathsf{a_0}$ to $\mathsf{a_1}$, from $\mathsf{a_1}$ to $\mathsf{a_2}$, and from $\mathsf{a_2}$ to $\mathsf{a_0}$ (see Fig.~\ref{fig:funnels}). The value of the barrier function threshold $\epsilon$ is set to be $-0.2$. In the barrier pair synthesis sub-problem, the value of barrier function decay rate $\alpha$ is set to be $1$. 
The video of the trace execution using $\mathsf{BP}$-$\mathsf{RRT}$ is available at \url{https://youtu.be/JiqQs1n9AM8}.

\section{Discussion}
 
In \eqref{eq:forward}, we define a forward kinematics equation with the same number of dimensions between the joint space and workspace. In the case of a redundant robotic system, the workspace position can be realized by an infinite number of joint space configurations in a manifold. Potential issues are raised in some of the barrier pair synthesis sub-problems, which rely on the unique solutions of the inverse kinematics function $\mathrm{R(\cdot)}$ in $\mathsf{LMI}$ \eqref{eq:x-inclusion}. To solve this type of issue, we can replace $\mathrm{R(\cdot)}$ by a pseudo-inverse of the Jacobian function $\mathrm{J(\cdot)}$ for enforcing uniqueness. 

In the $2$-$\mathsf{DOF}$ manipulator example, we set $\bar{x} = \bar{y} = 0.2 \, \mathrm{m}$ for defining the state-space constraints introduced in \eqref{eq:Q0}.
Under these constraints, the resulting volumes of the ellipsoidal regions of attraction are sufficient for covering the desirable regions and exploring the reachable workspace. 
The volumes of the ellipsoidal regions of attraction will be smaller if the barrier pair synthesis uses smaller values of $\bar{x}$ and $\bar{y}$.
However, if we adopt larger values of $\bar{x}$ and $\bar{y}$, the barrier pair synthesis does not guarantee to generate larger ellipsoidal regions of attraction due to the increment of the uncertainty in the norm-bound $\mathsf{LDI}$ model.
For achieving the optimal size of the ellipsoidal region of attraction, the barrier pair synthesis needs to keep a balance between the state-space constraints and the model uncertainty.

In this paper, the proposed $\mathsf{BP}$-$\mathsf{RRT}$ algorithm generates the low-level controllers for executing an accepting run of a nondeterministic Buchi automaton representing the given $\mathsf{LTL}$ specification.
In a more general case, the high-level discrete controller is in the form of a finite-state transition system instead of a particular accepting run of the $\mathsf{LTL}$ specification.
Similar to the process we show in our $2$-$\mathsf{DOF}$ manipulator example, the barrier pair sequences that execute the discrete state transitions can be created off-line using the $\mathsf{BP}$-$\mathsf{RRT}$ method and activated following the requests from the finite-state transition system.


\bibliographystyle{IEEEtran}
\bibliography{main}

\balance

\end{document}